\pdfoutput=1  % arXiv: build with pdflatex (PDF figures)
\documentclass[10pt,twocolumn,letterpaper]{article}

\usepackage[datasets]{wacv}

\usepackage{times}
\usepackage{epsfig}
\usepackage{graphicx}
\usepackage{amsmath}
\usepackage{amssymb}
\usepackage{booktabs}
\usepackage{multirow}
\usepackage{float}
\usepackage{capt-of}       %

\definecolor{wacvblue}{rgb}{0.21,0.49,0.74}
\usepackage[breaklinks,colorlinks,allcolors=wacvblue]{hyperref}

\graphicspath{{figs/}}

\makeatletter
\renewcommand\paragraph{\@startsection{paragraph}{4}{\z@}%
   {0.5ex}%
   {-1em}%
   {\normalfont\normalsize\bfseries}}
\makeatother
\newcommand{\numcells}{168}                 %
\newcommand{\numbackbones}{seven}
\newcommand{\numpixels}{$2.786\times10^{10}$}
\newcommand{\conf}{C}                       %
\newcommand{\claim}{\hat{\sigma}}           %

\begin{document}

\title{A Calibration Audit of Confidence in Feed-Forward 3D Reconstruction Models}

\author{
Nanxing Nick Deng$^{1,*}$
\and
Qing Cheng$^{1,2,*}$
\and
Niclas Zeller$^{3}$
\and
Daniel Cremers$^{1,2}$
\and
\\
$^1$Technical University of Munich \hspace{2em}
$^2$Munich Center for Machine Learning (MCML)
\\
$^3$Karlsruhe University of Applied Sciences
\\
$^{*}$Equal contribution.
}

\twocolumn[{%
  \renewcommand\twocolumn[1][]{#1}%
  \maketitle
  \begin{center}
    \includegraphics[width=0.88\textwidth]{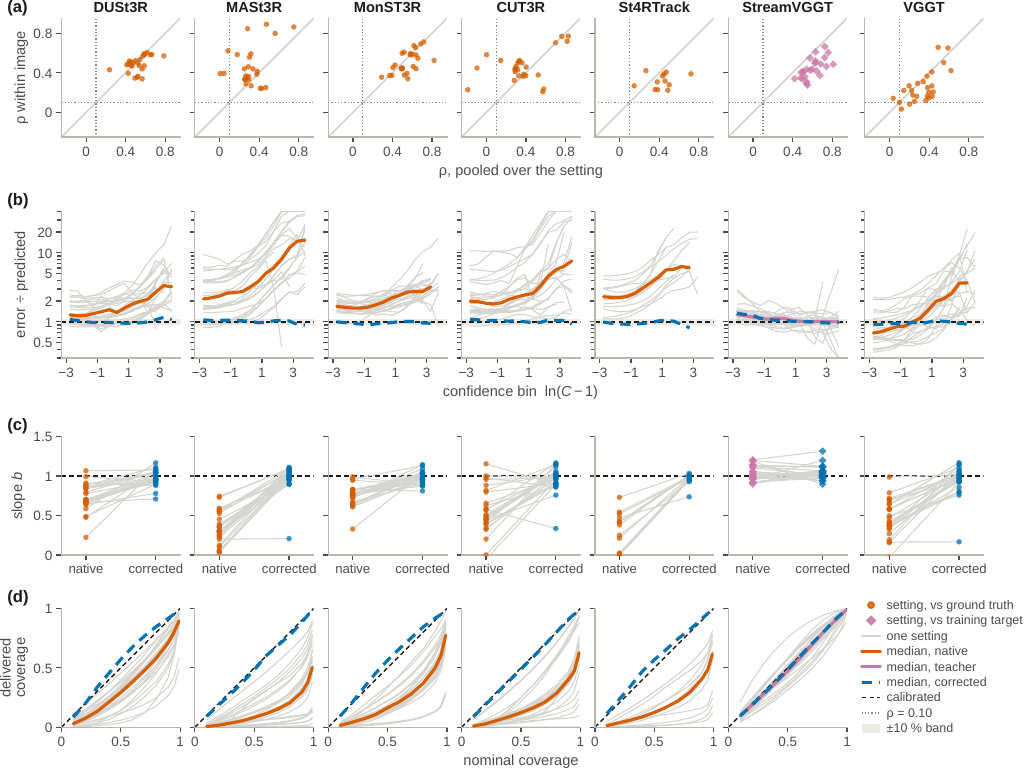}
    \captionof{figure}{The audit and the correction, over \numbackbones{}
    backbones (columns) and 13 datasets.
    \textbf{(a)}~Ranking, within an image against pooled.
    \textbf{(b)}~Level, mean error over predicted uncertainty by confidence bin.
    \textbf{(c)}~Slope of the calibration curve.
    \textbf{(d)}~Delivered against nominal coverage.
    Corrected curves are in-sample in (b) and (c), held out in (d). StreamVGGT
    is scored against its training teacher; VGGT's $\alpha$ is unattested, so
    its coverage is not shown.}
    \label{fig:audit}
  \end{center}
}]

\begin{abstract}
\looseness=-1
Feed-forward 3D reconstruction models output a per-pixel confidence that is used 
by downstream systems as an uncertainty signal. The confidence is trained to 
serve as a weight in the training loss of models. Whether the confidence can be used as an 
uncertainty magnitude has not been measured. We audit seven backbones on 13 datasets 
and score the confidence on four properties, i.e., ranking of error, ratio of error to 
uncertainty on average, slope of this ratio across the confidence range, 
and coverage of the implied error distribution.
Although the confidence ranks error quite well, the uncertainty decoded from the confidence 
is too small compared to the actual error. The uncertainty has the right size only under the 
exact training conditions. 
The median case is off by at least $2.4\times$ across all seven models, while the uncertainty 
is further off the more confident the model is. Our work shows that the overconfidence 
appears on unseen scenes even when the model reaches its loss's optimum. As a post-hoc 
repair we fit a power law on the confidence with two constants per backbone--dataset pair. 
The repair brings all four audited properties to target at the dataset level, while leaving 
ranking untouched. Fitted with the target dataset held out, the
constants bring the median case from $2.4\times$ off to $1.35\times$. The repair does
not hold below the dataset level, where two thirds of held-out scenes are still more
than five points off in coverage. We attribute what the repair cannot reach to the model, which 
carries neither the scale of the error nor the shape of its distribution across
predictions. We release the audit protocol, its results, and the fitted constants per
backbone-dataset pair.
\end{abstract}

\section{Introduction}
\label{sec:intro}

Feed-forward networks now regress dense 3D geometry from a set of images in a
single forward pass. Many of these networks predict, together with the
geometry, a per-pixel confidence that is trained as a weight on the regression
loss rather than as a predictive
uncertainty~\cite{wang2024dust3r,leroy2024mast3r,wang2025cut3r,wang2025vggt,zhang2025ffsurvey}.
However, the reconstruction systems built on them rely on the confidence to
indicate where the geometry is likely to be wrong, and use the confidence when
filtering and weighting the predictions. Several reconstruction pipelines
filter their output with a fixed confidence
threshold~\cite{wang2024dust3r,zhang2025monst3r,wang2025vggt,liu2025slam3r},
which assumes that a confidence value corresponds to the same error in every
prediction, since otherwise the accuracy of the retained points varies across
predictions. MP-SfM~\cite{pataki2025mpsfm} converts the confidence into the
standard deviation of a per-pixel depth prior during refinement, and that
standard deviation sets the weight on the prior term. If the standard
deviations are not proportional to the errors within a single prediction, the
pixels receive incorrect relative weights. To our knowledge, neither
assumption has been evaluated against the errors of these models.

A common approach to quantifying the predictive uncertainty of a network is to
represent that uncertainty as a probability distribution over the prediction.
Heteroscedastic regression~\cite{nix1994estimating,kendall2017uncertainties}
and evidential regression~\cite{amini2020deep} predict the parameters of that
distribution directly, while MC dropout~\cite{gal2016dropout} and deep
ensembles~\cite{lakshminarayanan2017simple} obtain the distribution from the
variation across repeated forward passes or ensemble members.
Trust3R~\cite{zhu2026trust3r} trains an evidential head for feed-forward
reconstruction and reports improved \emph{ranking}, a relative measure that
captures only the order the uncertainty assigns to the errors. However, a
threshold depends on the absolute magnitude of the uncertainty, which the
ranking does not constrain. Moreover, none of these methods derives a
distribution from the native confidence.

In this paper, we systematically evaluate the native confidence of seven
released backbones against the error that their predictions produce. We keep
the weights frozen and attach no additional head, so that the result describes
the confidence that deployed systems use. To interpret the confidence as an
uncertainty, we use the conversion implied by the optimum of the training
loss, a reciprocal of the confidence scaled by a constant of the loss.

We measure the ranking, level, slope and coverage of the resulting uncertainty
estimate against the error on every valid pixel. Level refers to the average
ratio of the error to the uncertainty estimate, while slope captures the rate
at which the error grows with the uncertainty estimate across the confidence
range. Coverage is the fraction of errors that fall inside the intervals the
uncertainty estimate implies, compared to the nominal rate. A fixed threshold
depends on the level, and the prior weight in MP-SfM depends on the slope. Our
evaluation spans 13 datasets and 168 combinations of backbone, dataset and
output head (Fig.~\ref{fig:audit}). We then compare post-hoc corrections and
measure how much of the miscalibration each removes without retraining.

We find that the confidence ranks the error well, while the level of the
uncertainty estimate is off by a median factor of $2.4$, with a larger factor
at higher confidence. Outside the conditions of training, the estimate is too
small, and the intervals derived from the estimate cover the error less often
than their nominal rate. Although a confidence head trained under the same
loss reaches the optimum of that loss on the scenes used for training, the
head remains overconfident on held-out scenes. The overconfidence therefore
cannot be explained by incomplete training. Instead, the confidence is optimal
only under the conditions of training, which are the target against which the
residual is measured, the normalization applied to the prediction, and the
data over which the average is taken. Once a model is deployed, any of the
three may change, and we attribute the overconfidence to that mismatch.

Among the corrections we compare, only a power law with two constants per
backbone--dataset pair corrects the level and the slope without altering the
ranking. A correction fitted per dataset generalizes to held-out scenes of
that dataset, and the best correction covers the error at the nominal rate
over the dataset as a whole. However, no correction we compare brings the
coverage of individual scenes to the nominal rate. The correction that
succeeds is fitted on each scene's own ground truth, which a deployed system
does not have.

\noindent\textbf{Contributions.}
\begin{itemize}\itemsep1pt \parskip0pt \topsep0pt \partopsep0pt
\item An evaluation protocol that interprets a native confidence as an uncertainty
through the conversion implied by the training loss and evaluates the result
in terms of ranking, level, slope and coverage. We apply the protocol to seven
released backbones on 13 datasets.
\item A convergence study showing that the overconfidence cannot be explained 
by incomplete training.
\item A comparison of the post-hoc corrections applicable on the frozen weights,
including existing methods, each evaluated on all four properties.
\item The two constants of the power law, fitted for every backbone--dataset pair
we evaluate. With the target dataset held out, the median case is off by a
factor of $1.35$.
\end{itemize}

\section{Related Work}
\label{sec:related}

\paragraph{Measures of dense uncertainty.}
Dense uncertainty is commonly judged by how well the predicted uncertainty
ranks the error. For optical flow, Ilg et al.~\cite{ilg2018uncertainty}
introduced the sparsification error curve and the area under that curve
(AUSE), and both were adopted for monocular depth~\cite{poggi2020uncertainty}.
Stereo confidence is evaluated with an ROC curve over the confidence
ranking~\cite{poggi2021confidence}. The same measures are used for uncertainty
estimated post hoc and without labels, on a frozen depth
network~\cite{hornauer2022gradient} or a fitted radiance
field~\cite{goli2024bayes}. MonoProb~\cite{marsal2024monoprob} and
SEDNet~\cite{chen2023learning} change the training objective to make the
prediction interpretable as an expected error, and each defines an absolute
metric. Gustafsson et al.~\cite{gustafsson2020evaluating} instead measure
interval coverage on dense depth, on heads trained in the same work. All
ranking measures are, however, invariant to a multiplicative error in the
predicted scale, and a model can achieve the best value on all of them while
understating the error by a constant factor.

\paragraph{Evaluations of the native confidence.}
The confidence we audit weights the residual during training. Kendall and
Gal~\cite{kendall2017uncertainties} refer to this weighting as learned loss
attenuation, and DUSt3R credits the particular form to
ConfNet~\cite{wan2018confnet}. Hillemann et
al.~\cite{hillemann2026uncertainty} evaluate that weight on VGGT, find the
depth branch overconfident, and conclude that metric uncertainty per predicted
3D point is not yet possible. Their evidence, however, comes from DTU alone
and from a monotonic, scale-free proxy they define. Since that proxy is
invariant to rescaling, their result establishes the direction of the defect
and not the size. In contrast, we measure that size on seven backbones and 13
datasets.

\paragraph{Replacing the head.}
Trust3R~\cite{zhu2026trust3r} trains a Normal-inverse-Wishart head on a frozen
MASt3R backbone, with the induced multivariate Student-$t$ as the predictive
distribution, and reports 25\,\% lower AURC and 41\,\% lower AUSE than
MASt3R's own confidence on ScanNet++. The two improvements are in ranking
metrics, measured on a retrained head over a pointmap the same work also
refines, and leave open whether a differently trained head is on average too
certain or too uncertain. We measure the same property on two formulations
whose uncertainty is not a loss weight, and find the uncertainty taken from
the disagreement among sampled predictions overconfident and the directly
regressed error underconfident.

\paragraph{Post-hoc recalibration.}
Correcting a frozen model's uncertainty against labelled data is standard
practice, by a sigmoid or a temperature on a classifier's
scores~\cite{platt2000probabilities,guo2017calibration} or by a monotone map
on a regressor's predictive distribution~\cite{kuleshov2018accurate}. For a
predicted standard deviation the usual correction is one scalar, fitted by
maximum likelihood on a labelled validation
set~\cite{levi2022evaluating,laves2021recalibration}. Conformal prediction
provides a finite-sample coverage guarantee for a correction of the same form,
either one constant for the calibration
set~\cite{papadopoulos2002inductive,lei2018distribution} or one per category
of a fixed partition~\cite{vovk2012conditional,bostrom2020mondrian}.
Refinements condition on overlapping groups~\cite{jung2023batch}, on kernel
neighbourhoods~\cite{guan2022localized} and on an estimated shift between
calibration and deployment~\cite{tibshirani2019conformal}. Other work
formulates the target as a risk~\cite{angelopoulos2025learn}. Such methods
have been applied to dense per-pixel
regression~\cite{angelopoulos2022imagetoimage} and to object pose estimation
from keypoints~\cite{yang2023object}, and for monocular depth Feldman et
al.~\cite{feldman2023achieving} construct per-pixel intervals under a rolling
risk-control guarantee. We are not aware of one for feed-forward pointmap
reconstruction. LoRA3D~\cite{lu2025lora3d} adapts the weights to a target
scene without labels, from the model's own multi-view predictions. We leave
the weights untouched and correct the confidence a released backbone already
emits.

\paragraph{Uncertainty in geometric systems.}
The uncertainty of a frozen geometric system has been corrected offline in
other contexts. Tsuei et al.~\cite{tsuei2021learned} map a visual-inertial
filter's covariance onto the empirical covariance with a network fitted on
held-out sequences. Correia Marques et
al.~\cite{correiamarques2024overconfidence} find semantic 3D mapping
overconfident and attribute the overconfidence in part to the independence
assumption of Bayesian fusion. MP-SfM~\cite{pataki2025mpsfm} rescales the
standard deviation of the depth prior by a constant tuned per estimator. The
transfer of such a constant beyond the tuning data has not been measured,
which we address in this work.
\section{Confidence Training}
\label{sec:loss}

\paragraph{Training Loss.}
The DUSt3R~\cite{wang2024dust3r} family of modelsis trained with a confidence-weighted 
regression loss and predicts a 3D point and a confidence
$\conf_i$ at every pixel $i$~\cite{leroy2024mast3r,zhang2025monst3r,wang2025cut3r,feng2025st4rtrack}.
The loss is defined as
\begin{equation}
\label{eq:confloss}
L = \sum_i \left( \conf_i\, e_i - \alpha \log \conf_i \right),
\end{equation}
where $\alpha>0$ is a fixed hyperparameter. $e_i$ is the error residual, which
the distance between the predicted 3D point and the target point. 
The first term of the loss weights the residual by its confidence, and the second term pushes 
the confidence up. The head does not emit the confidence as used in the loss, but instead 
it outputs $x$ such that $\conf = 1+e^{x}$. It follows that $\conf>1$. 

\paragraph{Point cloud normalization.}
Before the residual is computed during training, the predicted and the
ground-truth pointmap are each divided by a normalization factor. A pointmap's
factor is the mean distance of its points to the origin. The ground truth is
always divided by its own factor. The prediction is divided either by its own
factor or by the ground truth's. We call the first choice the scene-normalized 
branch and the second the metric branch.
Five of the backbones we audit use the prediction's 
own factor throughout. A prediction that is uniformly too large or too small is rescaled before the
residual is computed, and the model never learns absolute scale. MASt3R and
CUT3R use the ground truth's factor where the ground truth has a metric scale
and the prediction's own factor elsewhere. On those metric datasets a
prediction at the wrong overall size is penalized, and the model has to match
the scale of the ground truth.

\paragraph{Optimal Confidence.}
In order to determine whether the confidence can be used as an uncertainty estimate, first 
it needs to be resolved what the confidence is trained to represent. 
Setting $\partial L/\partial \conf_i = 0$ gives $\conf_i = \alpha/e_i$. 
Further, we know that the head of the model maps a per-pixel feature to a 3D point and a 
confidence. The head will map pixels to the same confidence, if the pixels features look alike 
in a certain way, while their residuals might differ. 
The loss over pixels of the same confidence depends on the mean of the residuals over these pixels, 
since $L$ is linear in $e$. The loss is minimized when the confidence equals $\alpha$ over that mean.
The confidence can be read as an uncertainty by inverting the trained relation,
\begin{equation}
\label{eq:decode}
\claim_i = \alpha / \conf_i ,
\end{equation}
which at the optimum returns $\mathbb{E}[e \mid \conf_i]$, the mean residual over
training pixels sharing pixel $i$'s confidence.
The mean residual depends on three conditions, which are the target the residual is
measured against, the normalization of the prediction, and the data the mean is
taken over.
With $z = e/\claim$, which is the residual in units of the predicted uncertainty,
the optimum becomes the statement
\begin{equation}
\label{eq:cond}
\mathbb{E}[z \mid \conf] = 1 .
\end{equation}

\section{Audit Protocol}
\label{sec:protocol}

\paragraph{Confidence to uncertainty.}
We convert the claimed uncertainty from confidence as $\claim = \alpha/\conf$,
which returns the mean residual over training pixels of the same confidence. The mean 
depends on the target training measured the residual against, the
normalization the prediction is divided by in training, and the data training saw.
 Wherever the release provides the
information we match the target and the normalization to training, leaving the
data as the condition that we alter. For the two branch-conditional backbones 
we keep one normalization convention across every setting, by scoring on the metric branch 
rather than applying a per-dataset rule.

\paragraph{Grid.}
\looseness=-1
Six of the \numbackbones{} backbones expose two output heads and one a single
head, giving $26\times6 + 12 = \numcells{}$ \emph{settings} over 13 datasets,
each scored on the four properties. The datasets are
ScanNet (val, train)~\cite{dai2017scannet}, TUM (static,
dynamic)~\cite{sturm2012benchmark}, 7-Scenes~\cite{shotton2013scene},
BlendedMVS~\cite{yao2020blendedmvs}, Bonn~\cite{palazzolo2019refusion},
KITTI~\cite{geiger2013kitti,uhrig2017sparsity}, CO3Dv2~\cite{reizenstein2021co3d},
DL3DV~\cite{ling2024dl3dv}, Sintel~\cite{butler2012sintel}, Virtual
KITTI~2~\cite{cabon2020vkitti2} and Dynamic
Replica~\cite{karaev2023dynamicstereo}, the same for every backbone and sampled
as eight-frame clips.
Backbones trained on some of these datasets are still scored on frames they did
not train on. Of the 13 datasets, 3 are close relatives of others, which leaves 10 
independent sources, and
one backbone cannot run CO3Dv2. Every setting is scored on its full pixel population, 
\numpixels{} pixels across
the grid. A pixel counts when its ground truth is valid, its depth is in range,
and its position is inside the model's field of view.

\paragraph{Rank correlation.}
Ranking is scored by the Spearman correlation $\rho$ between $\claim$ and $e$, and
a setting passes at $\rho \ge 0.10$. We report $\rho$ on two populations. The
\emph{pooled} correlation runs over every scored pixel of a setting. The
\emph{within-image} correlation is taken inside each image separately and
reported as the median over the setting's images. Measuring the correlation inside an image
rules out a confidence that only tracks how large the error from image to
image is. The within-image correlation excludes floored confidences and out-of-range bins
and runs on fewer pixels than the pooled one. The exclusion removes a median of
7\,\% of a setting's pixels and 81\,\% in the worst setting.

\paragraph{Confidence bins.}
Each pixel is assigned to one of 14 bins, and the bins are fixed on
$x = \ln(\conf-1)$, which inverts the head's activation $\conf = 1 + e^{x}$. The
$j$-th bin is $\mathcal{B}_j$ and holds $n_j$ pixels. Its calibration level is
\begin{equation}
\label{eq:ubar}
\bar{u}_j = \frac{1}{n_j}\sum_{i \in \mathcal{B}_j} z_i ,
\end{equation}
which is the ratio of the delivered error to the predicted uncertainty in that
bin. Eq.~\ref{eq:cond} requires $\bar{u}_j$ to equal one in every bin. We obtain
a confidence interval on $\bar{u}_j$ by resampling \emph{scenes} rather than
pixels. The pixels of one scene share a camera, a depth range and a
normalization, so they are not independent evidence. A bin fails high when its
interval lies entirely above $1.10$, and it fails low when the interval lies
entirely below $0.90$. We report a bin for a backbone only where at least half of 
the backbones' settings have pixels in the bin.

Each confidence bin has a predicted uncertainty $\claim_j$, taken at the bin
midpoint, and a mean delivered error $\bar{e}_j = \bar{u}_j\claim_j$. We fit
\begin{equation}
\label{eq:slope}
\ln \bar{e}_j = c + b\,\ln \claim_j
\end{equation}
across the bins of one setting, and we weight each bin by the number of pixels
that bin holds. The fit separates two defects. A factor that is common to every
bin is absorbed by the intercept $c$, so the exponent $b$ reports only how the
level changes from one bin to the next. Calibrated uncertainties give $c=0$ and
$b=1$, so doubling the predicted uncertainty doubles the delivered error. When
$b<1$, the predicted uncertainty varies more from one bin to the next than the
delivered error does.

\paragraph{Intervals.}
\looseness=-1
Coverage tests how well an error distribution with $\claim$ as scale covers
the actual error distribution. A correct $\claim$ puts the ground-truth point
within $T_p = \claim\cdot m_p$ of the predicted point for a fraction $p$ of the
pixels. The quantile $m_p$ is the level-$p$ quantile of a reference distribution
for $z = e/\claim$. We use two probability distributions for $z$. Read as a likelihood, 
the loss gives $z$ an exponential distribution, which is a Gamma distribution with shape
one. We fit the shape of the Gamma distribution. We also use
the empirically measured distribution of $z$ inside the pixel's own bin, which
assumes no family. Both distributions are fitted on a calibration half of a setting's scenes
and scored on the held-out half, at 11 levels from 10\,\% to 99\,\%. Because
the empirical distribution assumes no family, a shortfall under it means the
calibration scenes did not transfer to the held-out ones. For the coverage test, we read two numbers.
\emph{Delivered} coverage is the fraction of a scene's pixels whose ground-truth
point falls within $T_p$ of the predicted point. \emph{In band} is the fraction of a setting's held-out scenes whose own
delivered coverage falls within $0.05$ of the nominal level.
Sec.~\ref{sec:repair} reads both statistics over an eight-frame clip and over
a single frame as well, counting the fraction of held-out clips or frames in
band. 

\section{Calibration Audit}
\label{sec:audit}

\subsection{Results}

\paragraph{Ranking.}
Ranking is the one of the four properties we audit that holds.
Within images the median $\rho$ is $0.43$ and 166 of the 168 settings pass.
Pooled over a setting's pixels the median is $0.46$ and 161 pass
(Fig.~\ref{fig:audit}a, Tab.~\ref{tab:audit}). The seven settings that fail 
the pooled measurement lie on CO3Dv2 or BlendedMVS and pass within
images, meaning their confidence orders pixels inside an image while its scale does
not carry from one image to the next. In half the settings $\rho$ is negative 
on more than $8\,\%$ of their
images, and in the worst setting $\rho$ is negative on $44\,\%$ of the images.

\begin{figure*}[t]
\centering
\includegraphics[width=\textwidth]{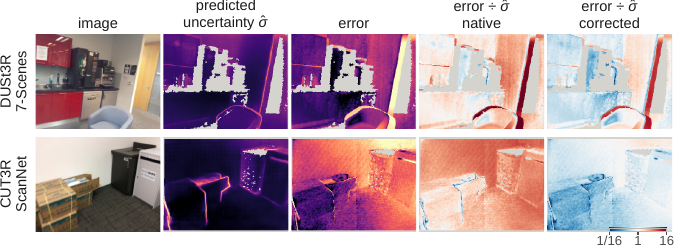}
\caption{One held-out frame at median \emph{level} from each of two settings.
The uncertainty follows the error's structure and sits below the error on 84 to
$97\,\%$ of the scored pixels. Our post-hoc correction removes most of the gap.}
\label{fig:frames}
\end{figure*}

\paragraph{Level.}
The predicted uncertainty is smaller than the error the model delivers. On the
five backbones trained against ground truth, 1{,}224 confidence bins fail and
every one of them is overconfident (Tab.~\ref{tab:audit},
Fig.~\ref{fig:frames}). The median setting is off by $2.4\times$ across the
grid and by $2.8\times$ on those five backbones. StreamVGGT is close to
calibrated when measured against the teacher it trained on and joins the others
when measured against ground truth. VGGT's hyperparameter $\alpha$ is unavailable and 
falls back to the default, therefore
VGGT's bins are reported and not aggregated into the total bin count.

\begin{table}[t]
\centering
\scriptsize
\setlength{\tabcolsep}{2.2pt}
\begin{tabular}{@{}l c cc cc c c c@{}}
\toprule
& & \multicolumn{2}{c}{Ranking} & \multicolumn{3}{c}{Level} & Slope & Coverage \\
\cmidrule(lr){3-4} \cmidrule(lr){5-7} \cmidrule(lr){8-8} \cmidrule(lr){9-9}
Backbone & \# & pooled & image & high & low & low $\to$ top bin & $b$ & cov \\
\emph{Target} & & \multicolumn{2}{c}{$\ge 0.10$} & 0 & 0 & $1 \to 1$ & $1$ & $0.90$ \\
\midrule

DUSt3R     & 26 & 0.52 & 0.49 & 208 & 0 & 1.26 $\to$ 3.22  & 0.70 & 0.71 \\
MonST3R    & 26 & 0.55 & 0.47 & 308 & 0 & 1.67 $\to$ 3.19  & 0.76 & 0.50 \\
MASt3R     & 26 & 0.29 & 0.39 & 293 & 0 & 2.14 $\to$ 15.24 & 0.30 & 0.29 \\
CUT3R      & 26 & 0.33 & 0.45 & 290 & 0 & 1.99 $\to$ 7.62  & 0.50 & 0.40 \\
St4RTrack  & 12 & 0.44 & 0.31 & 125 & 0 & 2.31 $\to$ 6.08  & 0.40 & 0.40 \\
\midrule
All five   & 116 & & & 1{,}224 & 0 & & & \\
\midrule
StreamVGGT & 26 & 0.59 & 0.43 & 40 & 10 & 1.26 $\to$ 1.01 & 1.01 & 0.90 \\
\quad vs.\ GT & & & & 297 & 0 & 2.47 $\to$ 3.35 & & \\
VGGT       & 26 & 0.37 & 0.21 & 120 & 64 & 0.69 $\to$ 3.67 & 0.42 & \\
\midrule
Grid       & 168 & 0.46 & 0.43 & & & & 0.65 & \\
\bottomrule
\end{tabular}
\caption{The audited properties with each entry being a median over the datasets. 
Ranking holds on every backbone. The level, the slope and the coverage
fail on every backbone read against ground truth, and the level worsens toward
the most confident bins. }
\label{tab:audit}
\end{table}

\paragraph{Slope.}
The audit shows that overconfidence grows with confidence. The slope $b$ lies 
entirely below one on 123 of the 168 settings (Fig.~\ref{fig:audit}c). The 
measured result is not an artifact of pooling the pixels of a setting's scenes, 
since the slope stays below one inside single frames.

\paragraph{Coverage.}
Our coverage test confirms that native confidence without correction can not construct an useful 
error distribution. The overconfidence of the level is reflected in the coverage. 
On the five backbones trained against ground truth the intervals are too narrow at all 11
nominal levels. In proportion the miss is worst at the low nominal levels. At
nominal $10\,\%$ no backbone delivers more than $3\,\%$
(Fig.~\ref{fig:audit}d).

\subsection{Candidate explanations}

\paragraph{Content and normalization.}
A backbone's level differs from dataset to dataset. We check whether the
content explains the variance. Splitting the datasets
by indoor against outdoor, static against dynamic, or real against synthetic
ground truth moves the level by at most $1.6\times$. Datasets outside every
training diet read no worse than those inside. The normalization training used
moves the level much more. CUT3R is $3.0\times$ off on the datasets it trained
on the metric branch and $15.6\times$ off on those it trained on the
scene-normalized branch. MASt3R moves the same way, from $12.6\times$ to
$31.8\times$. Neither model is calibrated on either branch.

\paragraph{Convergence.}
In no audited setting does the delivered error match the prediction in every
bin, which is what emitting the optimal confidence would require. Training may 
never have reached that optimum, or the target, the normalization
or the data may have changed between training and the audit.
 We check whether training converged on DUSt3R. A fresh confidence head
trained on DUSt3R's frozen trunk with two of DUSt3R's own datasets ends
calibrated on the frames used for training and overconfident on frames held
out. The released DUSt3R put back into training under the same loss behaves the
same way. Measured on CO3Dv2, a dataset in DUSt3R's diet, the level starts $2.79\times$ off,
reaches $1.10\times$ after roughly 300 updates, and stays between $0.91$ and
$1.12\times$ over 10 further epochs under both a cosine schedule from
$10^{-4}$ and a constant rate of $10^{-5}$. CO3Dv2 is in the training diet, and the 
scenes are likely ones training saw, but the
clips, crops and sampler are ours and not the release's, as well as the pixels
scored are not the exact pixels training consumed. The initial level measurement of 
$2.79\times$ therefore comes from conditions that differ from training. 
The loss reaches the optimal confidence on whatever data is currently being
trained on, and a longer run under the same loss would stop in the same place.
The target, the normalization and the data set the level. A few 
hundred updates under the same loss on labelled scenes of the target
domain would therefore repair the level there. Our runs used datasets from
DUSt3R's own diet, and we did not test an entirely new domain.

\paragraph{One condition at a time.}
We move each condition separately and leave the weights untouched. StreamVGGT,
scored against ground truth instead of the teacher that supervised training,
goes from calibrated to $2.2\times$. On ScanNet's training split, MASt3R's four
highest-confidence bins deliver $2.5$ to $5.9\times$ the claim on the metric
branch. The same bins deliver $1.1$ to $2.1\times$ on the scene-normalized
branch. The normalization therefore accounts for 55 to 64 percent of the overconfidence
on ScanNet, measured on the one backbone whose confidence can be read on both
branches. MonST3R and St4RTrack read $1.38$ and $0.98$
on the scenes they trained on and $2.34$ and $6.05$ on unseen scenes of the
same datasets.

\paragraph{Ground-truth quality.}
DUSt3R is no less overconfident on synthetic datasets whose rendered ground
truth is exact. The residual we score also includes any error carried by the
ground truth. We allow a conservative amount of measurement noise, which covers
at most $12.7\,\%$ of the failing bins. Ground-truth quality does not explain
the overconfidence.

\subsection{Other formulations}
The level fails under other designs as well, and they fail in both directions.
Marigold~\cite{ke2024repurposing,ke2025marigold} has no trained confidence head
and takes uncertainty from the disagreement among 10 sampled predictions. 
The result is overconfident on all 13 datasets. The nominal $90\,\%$ interval only 
covers $17$ to $41\,\%$ of the pixels.
UniDepth~v2~\cite{piccinelli2026unidepthv2} regresses the error directly and is
underconfident instead. Only the 10 datasets with metric ground truth can score
a prediction in metres, and on those the prediction is 3 to 34
times the error the model delivers.

\section{Post-hoc Recalibration}
\label{sec:repair}

\newcommand{\bst}[1]{\underline{\textbf{#1}}}
\begin{table*}[t]
\centering
\footnotesize
\setlength{\tabcolsep}{3.6pt}
\begin{tabular}{@{}l c c c c c c c c c c@{}}
\toprule
& & \multicolumn{2}{c}{Ranking} & \multicolumn{4}{c}{Level} & Slope & \multicolumn{2}{c}{Coverage} \\
\cmidrule(lr){3-4} \cmidrule(lr){5-8} \cmidrule(lr){9-9} \cmidrule(lr){10-11}
Correction & par. & pooled & image & overall & low bin & top bin & spread & $b$ & cov & in band \\
\emph{Target} & & $\uparrow$ & $\uparrow$ & 1 & 1 & 1 & 0 & 1 & 0.90 & $\uparrow$ \\
\midrule
\multicolumn{11}{@{}l}{\emph{Rescaling the predicted uncertainty}} \\
none, native & 0 & 0.46 & 0.428 & $2.41\times$ & 1.48 & 3.60 & 0.258 & 0.65 & 0.570 & 0.000 \\
one scalar per model & 1 & $=$ & $=$ & $1.41\times$ & 0.56 & 1.33 & 0.258 & $=$ & 0.917 & 0.133 \\
one scalar per dataset~\cite{papadopoulos2002inductive,guo2017calibration,laves2021recalibration}
 & 1 & $=$ & $=$ & $1.14\times$ & 0.59 & 1.47 & 0.258 & $=$ & \bst{0.898} & 0.258 \\
\quad fitted on the other twelve & 1 & $=$ & $=$ & $1.46\times$ & 0.55 & 1.37 & 0.258 & $=$ & 0.914 & 0.106 \\
one per confidence bin~\cite{vovk2012conditional,bostrom2020mondrian}
 & 14 & $=$ & 0.376 & \bst{$1.11\times$} & \bst{0.98} & \bst{1.07} & 0.144 & \bst{1.00} & \bst{0.902} & \bst{0.333} \\
\quad rank-preserving & $\le$14 & $=$ & 0.428 & $1.14\times$ & 0.59 & 1.62 & 0.259 & 0.65 & 0.888 & 0.276 \\
\quad power law, two constants & 2 & $=$ & $=$ & \bst{$1.11\times$} & 1.03 & 0.92 & \bst{0.125} & 0.99 & 0.905 & \bst{0.333} \\
\midrule
\multicolumn{11}{@{}l}{\emph{Re-choosing the interval, leaving the predicted uncertainty untouched}} \\
isotonic~\cite{kuleshov2018accurate} & 1 & $=$ & $=$ & $=$ & $=$ & $=$ & $=$ & $=$ & 0.893 & 0.267 \\
four published refinements, vs.\ per-bin & 7--70 & $=$ & $=$ & $=$ & $=$ & $=$ & $=$ & $=$ & --- & $+0.000$ \\
\midrule
\multicolumn{11}{@{}l}{\emph{Fitted on the scenes it is scored on, so a ceiling rather than a correction}} \\
one scalar per scene & 1/scene & --- & $=$ & $1.00\times$ & 0.69 & 1.42 & 0.198 & 0.72 & 0.914 & 0.867 \\
\bottomrule
\end{tabular}
\caption{Post-hoc corrections on all four properties, as medians over the
\numcells{} settings, fitted and scored on disjoint halves of a dataset's
scenes. Subtracting two rows does not give the change on a single setting. Only
the power law repairs the level and the slope while leaving the ranking
untouched. Each in-band value is a fraction of a median of nine held-out scenes
per setting. ``$=$'' marks a cell unchanged from the native row.}
\label{tab:unified}
\end{table*}

\paragraph{Correction types.}
We compare the corrections that can be applied to a frozen model. We score each
correction on all four properties (Tab.~\ref{tab:unified}), because repairing one
property can damage another. A rescaling correction multiplies the native
uncertainty by a factor that depends on the confidence value and the dataset,
and on no other property of the pixel. Isotonic recalibration and four conformal
refinements change no uncertainty at all, and re-choose the interval the native
uncertainty implies. Among the rescaling corrections, $\sigma$-scaling holds the
factor constant across the confidence range by applying one value per backbone--dataset
pair~\cite{laves2021recalibration}. A per-bin fit lets the factor vary freely
from one confidence bin to the next, as in Mondrian conformal
prediction~\cite{vovk2012conditional,bostrom2020mondrian}.

\paragraph{Power-law correction.}
We propose a rescaling correction whose factor varies with confidence along a
line. Eq.~\ref{eq:slope} already fits the line, with the audited slope as the
exponent, and we take the error the line predicts as the corrected uncertainty,
\begin{equation}
\label{eq:power}
\claim' = e^{c}\,\claim^{\,b}
\qquad\text{that is,}\qquad
g(\claim) = e^{c}\,\claim^{\,b-1}.
\end{equation}
Refitting Eq.~\ref{eq:slope} on the corrected uncertainty returns $c=0$ and
$b=1$ by construction. We fit the pair as
$\claim' = \claim\exp(a + s\ln(\claim/\alpha))$ with $s = b-1$ and
$a = c + s\ln\alpha$, where $\alpha$ is the loss constant. The constant $a$ sets
the level and $1{+}s$ sets the slope. We clamp the exponent at $0.05$, because a
non-positive exponent would make $\claim \mapsto e^{c}\claim^{\,b}$
non-increasing and flip the confidence ordering. The fit falls below $0.05$ on
10 settings, and the exponent is non-positive on 6 of the 10. The clamp leaves
the ranking unchanged on all 168 settings.

We fit the two constants on half of a dataset's scenes and score them on the
other half. On scenes the fit never saw, the level and the slope return to
target (Tab.~\ref{tab:unified}). By comparison, a separate constant in every bin
reaches the same level, but the constant jumps at each bin boundary. Two pixels
on either side of a boundary can then swap order, costing each setting a median
$0.0138$ of within-image $\rho$ against the native value. Forcing the per-bin
fit to preserve the order restores the ranking, but the slope falls back to the
native value (Tab.~\ref{tab:unified}, rank-preserving row). Only the power law
repairs the level and the slope while leaving the ranking untouched.

Tab.~\ref{tab:constants} lists the two power-law constants for every
backbone--dataset pair. We also provide two constants per backbone, the median
over that backbone's settings, that can be applied to an unseen dataset. We test
the per-backbone constants by holding out one of the 13 datasets at a time and
taking the median over the other twelve. The level then reaches a grid median
$1.35\times$ and moves closer to one on $74\,\%$ of settings. Although a close
relative of the target dataset can remain among the other twelve, removing the
relatives as well leaves both numbers unchanged. One scalar per dataset, held
out the same way, ends further from one (Tab.~\ref{tab:unified}, fitted on the
other twelve).

\begin{table*}[t]
\renewcommand{\arraystretch}{0.88}%
\centering
\scriptsize
\setlength{\tabcolsep}{4pt}
\begin{tabular}{@{}l *{7}{cc}@{}}
\toprule
Dataset & \multicolumn{2}{c}{DUSt3R} & \multicolumn{2}{c}{MonST3R} & \multicolumn{2}{c}{MASt3R} & \multicolumn{2}{c}{CUT3R} & \multicolumn{2}{c}{St4RTrack} & \multicolumn{2}{c}{StreamVGGT} & \multicolumn{2}{c}{VGGT} \\
\cmidrule(lr){2-3} \cmidrule(lr){4-5} \cmidrule(lr){6-7} \cmidrule(lr){8-9} \cmidrule(lr){10-11} \cmidrule(lr){12-13} \cmidrule(lr){14-15}
 & $a$ & $1{+}s$ & $a$ & $1{+}s$ & $a$ & $1{+}s$ & $a$ & $1{+}s$ & $a$ & $1{+}s$ & $a$ & $1{+}s$ & $a$ & $1{+}s$ \\
\midrule
ScanNet & $-0.36$ & 0.70 & $+0.21$ & 0.81 & $-0.09$ & 0.38 & $-0.12$ & 0.35 & $+0.07$ & 0.32 & $-0.26$ & 0.90 & $-0.82$ & 0.63 \\
ScanNet-tr & $-0.42$ & 0.72 & $+0.22$ & 0.77 & $-0.13$ & 0.35 & $+0.04$ & 0.45 & $+0.15$ & 0.37 & $-0.18$ & 0.94 & $-0.95$ & 0.62 \\
7-Scenes & $+0.12$ & 0.72 & $+0.75$ & 0.84 & $+0.24$ & 0.38 & $+0.38$ & 0.57 & $+0.59$ & 0.23 & $-0.09$ & 0.92 & $-0.24$ & 0.45 \\
TUM & $+0.06$ & 0.78 & $+0.59$ & 0.81 & $+0.29$ & 0.19 & $+0.59$ & 0.47 & $+0.88$ & 0.56 & $-0.25$ & 0.93 & $-0.09$ & 0.52 \\
TUM-dyn & $+0.73$ & 1.03 & $+0.94$ & 0.97 & $+0.76$ & 0.65 & $+1.12$ & 1.08 & $+0.46$ & 0.73 & $+0.28$ & 1.09 & $-0.37$ & 0.52$^\dagger$ \\
Bonn & $-0.07$ & 0.41 & $+0.39$ & 0.55 & $+0.16$ & 0.34 & $+0.54$ & 0.61 & $+1.30$ & 0.42 & $-0.03$ & 1.01 & $-0.31$ & 0.37 \\
DL3DV & $+0.38$ & 0.51 & $+0.48$ & 0.65 & $+1.44$ & 0.34 & $+1.60$ & 0.41 & $+1.55$ & 0.42 & $+0.04$ & 1.04 & $+0.96$ & 0.57 \\
CO3Dv2 & $+0.12$ & 0.86 & $+0.09$ & 0.77 & $+1.41$ & 0.05$^\dagger$ & $+1.27$ & 0.05$^\dagger$ & \multicolumn{2}{c}{--} & $+0.25$ & 1.01 & $+0.40$ & 0.39 \\
BlendedMVS & $+0.44$ & 0.69 & $+0.47$ & 0.84 & $+1.33$ & 0.07$^\dagger$ & $+1.50$ & 0.21$^\dagger$ & $+1.03$ & 0.55 & $+0.19$ & 1.06 & $-1.03$ & 0.08$^\dagger$ \\
KITTI & $+0.79$ & 0.91 & $+0.73$ & 0.89 & $+1.46$ & 0.40 & $+0.56$ & 1.07 & $+0.72$ & 0.05$^\dagger$ & $+0.21$ & 1.02 & $+0.44$ & 0.74 \\
VKITTI2 & $+0.70$ & 0.83 & $+0.61$ & 0.68 & $+1.60$ & 0.50 & $+0.01$ & 0.93 & $+1.16$ & 0.06 & $+0.18$ & 1.07 & $-0.42$ & 0.48 \\
Sintel & $+1.02$ & 0.77 & $+0.50$ & 0.74 & $+0.79$ & 0.54 & $+0.77$ & 0.96 & $+1.05$ & 0.05$^\dagger$ & $-0.02$ & 0.92 & $+0.51$ & 0.68 \\
Dyn.Replica & $-0.07$ & 0.85 & $+0.28$ & 0.62 & $-0.31$ & 0.61 & $+0.02$ & 0.63 & $+0.41$ & 0.70 & $-0.28$ & 1.01 & $-0.81$ & 0.59 \\
\midrule
\emph{median over settings} & $+0.19$ & 0.78 & $+0.42$ & 0.79 & $+0.70$ & 0.38 & $+0.62$ & 0.49 & $+0.80$ & 0.39 & $-0.04$ & 0.98 & $-0.22$ & 0.51 \\
\midrule
level, native & \multicolumn{2}{c}{1.78} & \multicolumn{2}{c}{2.54} & \multicolumn{2}{c}{5.65} & \multicolumn{2}{c}{3.39} & \multicolumn{2}{c}{3.00} & \multicolumn{2}{c}{1.11} & \multicolumn{2}{c}{1.90} \\
\quad with the median pair, dataset held out & \multicolumn{2}{c}{1.42} & \multicolumn{2}{c}{1.24} & \multicolumn{2}{c}{2.22} & \multicolumn{2}{c}{1.74} & \multicolumn{2}{c}{1.58} & \multicolumn{2}{c}{1.13} & \multicolumn{2}{c}{1.52} \\
\quad with a refit pair, a few target scenes & \multicolumn{2}{c}{1.12} & \multicolumn{2}{c}{1.14} & \multicolumn{2}{c}{1.18} & \multicolumn{2}{c}{1.13} & \multicolumn{2}{c}{1.06} & \multicolumn{2}{c}{1.06} & \multicolumn{2}{c}{1.12} \\
\bottomrule
\end{tabular}
\caption{The released correction of the native uncertainty, one per
backbone--dataset pair, applied as $\claim' = \claim\exp(a +
s\ln(\claim/\alpha))$. A dagger marks a pair with at least one head's exponent
clamped at $0.05$.}
\label{tab:constants}
\end{table*}

\paragraph{Choice of correction.}
The power law and the per-bin fit deliver the same level, slope and coverage,
and only the power law keeps the ranking. Both corrections cover the error at
the nominal $90\,\%$ over a dataset as a whole (Tab.~\ref{tab:unified}, cov
column), and both put a median $\approx 33\,\%$ of a setting's held-out scenes
in band (in-band column). No correction we score on held-out scenes of the same
dataset raises the fraction, isotonic recalibration included. We recommend the
power law wherever the uncertainty weights a residual across the confidence
range~\cite{jung2025uncertainty}.

\paragraph{Limit of rescaling.}
We further show the limits of correcting the coverage of the native uncertainty.
Coverage is scored through an error distribution fitted on half of a dataset's
scenes, and the distribution does not transfer below the dataset. We fit a
scalar on each held-out scene, keep the error distribution fitted on the
dataset's calibration scenes, and score on the same scenes. We observe an
in-band fraction of $\approx 87\,\%$ (Tab.~\ref{tab:unified}), a median gain of
42 percentage points per setting and positive on 148 of the 168 settings. We
then replace the scalar by the scene's own error quantiles, at the 11 nominal
levels and on the same labels, and every held-out scene falls in band by
construction. A scalar is therefore not enough to repair coverage at the scene
level, because scenes differ in the shape of their error distribution. The same
effect appears below the scene level. Counted per eight-frame clip and per
frame, the scene-fitted quantiles put $41\,\%$ of clips and $26\,\%$ of frames
in band. A deployed system has no labels for the scene at hand, and no
correction we score without labels reaches the $\approx 87\,\%$ ceiling.

\paragraph{Reweighting calibration frames.}
We test whether four published refinements fix the coverage failure of held-out
scenes~\cite{vovk2012conditional,jung2023batch,guan2022localized,tibshirani2019conformal}.
Each builds the interval for a held-out frame from the calibration frames, and
gives more weight to the calibration frames that resemble the held-out frame.
Resemblance is measured on how many of a frame's pixels fall in each confidence
bin, which the model gives without ground truth. No refinement raises the in-band
fraction above treating every calibration frame alike, and each improves 15 to
$36\,\%$ of settings while degrading a comparable share
(Tab.~\ref{tab:unified}). The resemblance the refinements can see is not the
resemblance that matters. Held-out scenes fail coverage because the error level
at a given confidence differs from one scene to the next, and finding the
calibration frames that share a held-out scene's error level would need labels
from that scene. We could not settle whether the error level follows observable
properties of a scene.

\paragraph{Corrections with additional input.}
Two corrections outside Tab.~\ref{tab:unified} read more than the confidence
value. Only MASt3R lets us divide the predicted uncertainty by the frame's own
predicted scale. Over 14 metric-branch settings the division closes a median
half of the gap between the per-bin correction and the per-scene ceiling. We
also fit a linear model on the backbone's frozen features to predict a corrected
uncertainty for each pixel. On 6 of 9 settings of three backbones, the model
brings the slope a median $94\,\%$ of the way to one on held-out scenes. Neither
correction improves the ranking.

\section{Conclusion}
\label{sec:conclusion}

We audited the per-pixel confidence of \numbackbones{} released feed-forward
reconstruction models on 13 datasets, read as the uncertainty the training loss
implies. We find that the confidence orders the error inside an image on every
backbone, that the models are overconfident wherever ground truth allows the
comparison, and that the overconfidence is largest where the models are most
confident. Training longer does not fix the
overconfidence. We drive a confidence head to the optimum of the training loss
and find the head calibrated on training frames and overconfident on held-out
frames. The optimum is defined against the target, the
normalization and the data of training, and a deployment can change each
condition. We repair most of the overconfidence with a power law of two
constants per backbone and dataset, and the correction leaves the ranking
untouched. We release the constants, the protocol and the underlying
measurements. The correction is right for a dataset as a whole and not for single
scenes, because scenes differ in how widely errors spread, and one pair
of constants cannot follow scene-to-scene variation. We could not predict the
spread of a scene without labels from the scene, and we leave the question open.

{\small
\bibliographystyle{ieeenat_fullname}
\bibliography{refs}
}

\end{document}